\pdfoutput=1
\documentclass[11pt]{article}

\usepackage{acl}

\usepackage{times}
\usepackage{latexsym}
\usepackage[T1]{fontenc}
\usepackage[utf8]{inputenc}
\usepackage{microtype}
\usepackage{inconsolata}
\usepackage{graphicx}
\usepackage{amsmath}
\usepackage{mathtools}
\usepackage{tikz}
\usepackage{url}
\usepackage[edges]{forest}
\usepackage{graphicx}         
\usepackage{natbib}           
\usepackage{lipsum}           
\definecolor{hidden-draw}{RGB}{20,68,106}
\definecolor{hidden-pink}{RGB}{255,245,247}
\title{A Survey on Parallel Reasoning}

\author{Ziqi Wang$^{1,2*}$, Boye Niu$^{2,3*}$, Zipeng Gao$^{1*}$, Zhi Zheng$^1$, Tong Xu$^1$ \\
\textbf{Linghui Meng$^2$, Zhongli Li$^2$, Jing Liu$^2$, Yilong Chen$^2$, Chen Zhu$^1$}\\
\textbf{Hua Wu$^2$, Haifeng Wang$^{2\dag}$, Enhong Chen$^{1\dag}$}\\
\\
$^1$USTC, $^2$Baidu, $^3$USYD \\
$^*$Equal Contributors, $^{\dag}$Corresponding Authors \\
\texttt{\{wzq142857, gaozp619\}@mail.ustc.edu.cn, bniu6645@uni.sydney.edu.au}
}

\begin{document}
\maketitle
\begin{abstract}
With the increasing capabilities of Large Language Models (LLMs), parallel reasoning has emerged as a new inference paradigm that enhances reasoning robustness by concurrently exploring multiple lines of thought before converging on a final answer. 
It has become a significant trend to explore parallel reasoning to overcome the fragility of standard sequential methods and improve practical performance. 
In this paper, we aim to survey and summarize the progress and challenges of parallel reasoning. 
We first present a formal definition of parallel reasoning and clarify its distinction from related concepts like Chain-of-Thought.
Then, we organize and discuss advanced techniques based on a novel taxonomy, including non-interactive reasoning, interactive reasoning, and efficiency-focused decoding strategies. Additionally, we explore various application scenarios, such as solving complex problems and enhancing the reliability of LLM outputs. 
Finally, we highlight the core challenges of parallel reasoning and suggest potential directions for future research. 
We hope that our work can provide a useful roadmap for beginners and encourage more research on improving parallel reasoning methods.
Related source can be avaliable in \url{https://github.com/PPPP-kaqiu/Awesome-Parallel-Reasoning}.
\end{abstract}

\section{Introduction}
Modern large language models (LLMs) have acquired powerful foundational capabilities by scaling parameters and training data during their development~\citep{gpt3, palm, gpt4, llama, llama2, KIMI2}. 
\begin{figure}[htbp]
    \centering
    \includegraphics[width=.45\textwidth]{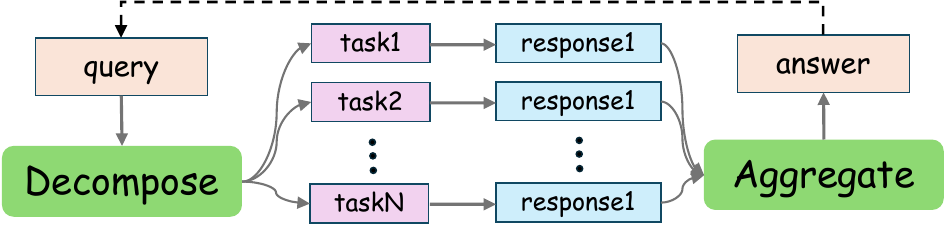}
    \caption{An overall framework for a recurrent parallel reasoning paradigm. The process begins with a decomposition stage, followed by parallel processing of sub-tasks, and concludes by aggregating the outputs into a single answer.}
    \label{fig1}
\end{figure}
Subsequent efforts explored inference-time scaling that extends the Chain-of-Thought(CoT), demonstrating significant improvements in reasoning performance~\citep{CoT, gpt-o1, deepseekR1}. 
Building on these successes, we examine an orthogonal approach: concurrently exploring multiple reasoning paths to broaden inference and further enhance reasoning depth and breadth.

Standard sequential reasoning often proves fragile on complex tasks, as it can fall into the so-called “prefix trap”~\citep{Learning-from-Peers}; once the model commits to an early reasoning path, it struggles to self-correct and may never reach the optimal solution. 
This weakness is starkly illustrated by the gap between single-pass performance (Pass@1) and the best outcomes achieved through multiple samples (Pass@k), showing that a sequential reasoning approach cannot fully leverage the model’s potential~\citep{EvaluatingLargeLanguageModels}. In contrast, parallel reasoning mimics a breadth-first search by exploring multiple paths at once, boosting robustness and tapping into a wider range of the model’s problem-solving abilities.

The key idea of parallel reasoning is to make multiple attempts in parallel before answering a question and then aggregate the proposed solutions. 
Figure~\ref{fig1} provides an example of this paradigm. 
The process begins with a divergence step, wherein the language model concurrently explores multiple reasoning paths. 
This is achieved either by generating several complete solution candidates in parallel or by decomposing the problem into sub-tasks that are solved simultaneously.
Following this parallel generation, a convergence step aggregates the various outputs to produce a single, final response. 
In contrast to the serial execution of a single reasoning path typical of sequential reasoning, the parallel method is designed to enhance robustness by identifying the correct answer from a diverse set of potential solutions.

As a novel inference paradigm, parallel reasoning (PR) offers several compelling advantages. First, by exploring multiple reasoning paths before converging on a final answer, PR can produce higher-quality responses than single-path methods — thereby enhancing user experience in real-world applications. 
This represents a genuine improvement in the model's practical performance, distinguishing it from metrics like pass@K, which reflect theoretical potential but, being ground-truth-based, do not measure the performance a user actually experiences.
Second, PR is orthogonal to the chain-of-thought paradigm~\citep{CoT, Verify-Step-by-Step, Complexity-Based-Prompting-for-Multi-step-Reasoning}: whereas CoT extends reasoning depth through step-by-step inference, PR expands reasoning breadth via concurrent exploration. This design enables PR to scale independently and continue benefiting from advances in foundation model capabilities.
Third, PR can achieve comparable or even superior performance to sequential reasoning with improved computational efficiency, especially when paired with algorithmic-system co-design optimizations such as KV cache reuse~\citep{Multiverse}. 
Furthermore, the core philosophy of parallelism embodied by PR also informs broader strategies for accelerating inference, including techniques like parallel decoding and speculative execution.

To structure this landscape, Figure~\ref{taxo} presents a taxonomy of parallel reasoning along three key dimensions:
(1). Non-interactive Parallel Reasoning: The model independently generates various attempts and then produces a final reply.
(2). Interactive Parallel Reasoning: This involves active interaction within the reasoning paths during the inference process.
(3). Efficiency: This covers both engineering designs aimed at accelerating parallel reasoning and novel decoding methods guided by the principles of parallelism.
We highlight the challenges and potential directions, and hope our work provides a useful road map for beginners interested in this area and sheds light on future research.
\tikzstyle{my-box}=[
    rectangle,
    draw=hidden-draw,
    rounded corners,
    text opacity=1,
    minimum height=1.5em,
    minimum width=5em,
    inner sep=2pt,
    align=center,
    fill opacity=.5,
    line width=0.8pt,
]
\tikzstyle{leaf}=[my-box, minimum height=1.5em,
    fill=hidden-pink!80, text=black, align=left,font=\normalsize,
    inner xsep=2pt,
    inner ysep=4pt,
    line width=0.8pt,
]
\begin{figure*}[t!]
    \vspace{-1.0cm}
    \centering
    \resizebox{\textwidth}{!}{
        \begin{forest}
            forked edges,
            for tree={
                grow=east,
                reversed=true,
                anchor=base west,
                parent anchor=east,
                child anchor=west,
                base=left,
                font=\large,
                rectangle,
                draw=hidden-draw,
                rounded corners,
                align=left,
                minimum width=4em,
                edge+={darkgray, line width=1pt},
                s sep=3pt,
                inner xsep=2pt,
                inner ysep=3pt,
                line width=0.8pt,
                ver/.style={
                    rotate=90, 
                    child anchor=north, 
                    parent anchor=south, 
                    anchor=center
                },
            },
            where level=1{text width=6.5em,font=\normalsize,}{},
            where level=2{text width=12.5em,font=\normalsize,}{},
            where level=3{text width=10.0em,font=\normalsize,}{},
            where level=4{text width=6.0em,font=\normalsize,}{},
            [
                Parallel Reasoning, ver
                [
                    Non-interactive, ver
                    [
                        Self-Consistency (\S \ref{Self-Consistency})
                        [
                         Self-Consistency~\cite{Self-Consistency}{, }Adaptive-Consistency~\cite{Let's-Sample-Step-by-Step}{,}DeepConf~\cite{DeepConf}\\ 
                         CWSC~\cite{CWSC}{, }SOFT-SC~\cite{Soft-Self-Consistency}{, }USC~\cite{Universal-self-consistency}, leaf, text width=48.5em
                        ]
                    ]
                    [
                        Ranking base (\S \ref{ranking})
                        [   
                            Best-of-N Sampling
                            [
                            Compute‑Optimal Scaling~\cite{Scaling-LLM-Test-Time-Compute}{, }LLM-Monkeys~\cite{Large-Language-Monkeys}, leaf, text width=37.0em
                            ]
                        ]
                        [   
                            Advancing Verifier
                            [
                            Training-Verifiers~\cite{Training-Verifiers-to-Solve-Math-Word-Problems}{, }Process\&Outcome~\cite{Solving-math-word-problems-withprocess-and-outcome-based-feedback}{, }\\MATH-SHEPHERD~\cite{Math-Shepherd}{, }Generative-Verifiers~\cite{Generative-Verifiers}{, }\\Latency-TTC~\cite{latent-ttc}, leaf, text width=37.0em
                            ]
                        ]
                        [   
                              Ranking Mechanism
                              [
                              BoNBoN~\cite{BoNBoN-Alignment}{, }PairJudge-RM~\cite{PairJudge-RM}, leaf, text width=37.0em
                              ]
                        ]
                        [   
                              Generative Synthesis
                              [
                              A2R~\cite{A2R}{,}SSA~\cite{Learning-to-Reason-Across-Parallel}{, }\\GSR~\cite{learn-refine}{, }SEAT~\cite{Adaptive-Termination-for-Multi-round-Parallel}, leaf, text width=37.0em
                              ]
                        ]
                    ]
                    [
                        Structure reasoning (\S \ref{structure})
                        [
                            Foundational Structures
                            [
                            GoT~\cite{Graph-of-Thoughts}{, }ToT~\cite{Tree-of-Thoughts}{, }CR~\cite{Cumulative-Reasoning-with-Large-Language-Models}, leaf, text width=37.0em
                            ]
                        ]
                        [
                            Guided Search \&\\ Efficiency Optimization
                            [
                            Self-Evaluation~\cite{Self-Evaluation}{, }Supervision with MCTS~\cite{Enhancing-Reasoning-through-process-supervision}{, }\\SoT~\cite{Skeleton-of-Thought}{, }DPTS~\cite{Dynamic-Parallel-Tree-Search-for-Efficient-LLM-Reasoning}{, }AoT~\cite{Atom-of-Thoughts}, leaf, text width=37.0em
                            ]
                        ]
                        [
                            Other Paradigm
                            [
                            Self-ask~\cite{Measuring-and-Narrowing}{, }ThinkSum~\cite{ThinkSum}{,}RAP~\cite{Reasoning-with-Language-Model}, leaf, text width=37.0em
                            ]
                        ]
                    ]
                ]    
                [
                    Interactive, ver
                    [
                        Intra-interaction (\S \ref{subsec:intra})
                        [   
                        Leap~\cite{Learning-from-Peers}{, }Hogwild!~\cite{Hogwild}{, }APR~\cite{Learning-Adaptive-ParallelReasoning}{,}Parallel-R1~\cite{parallel-r1}{, }\\HDS~\cite{hds}{,}MIRAGE~\cite{MIRAGE}{,}Group-think~\cite{Group-Think}{, }CoSD~\cite{Speculatethen}, leaf, text width=48.5em
                        ]
                    ]
                    [
                        Inter-interaction (\S \ref{subsec:inter})
                        [
                            Debate \\ Reflection
                            [  
                            Multiagent Debate~\cite{ImprovingFactualityandReasoning}{, }RECONCILE~\cite{ReConcile}{, }\\RR-MP~\cite{EnhancingLLMReasoningwithMulti-Path}{, }ChatEval~\cite{chateval}, leaf, text width=37.0em
                            ]
                        ]
                        [
                            Collaboration \\ Division of Labor
                            [
                            CoA~\cite{ChainofAgents}{, }Corex~\cite{Corex}{, }CoMM~\cite{CoMM}{, }\\DoT~\cite{Division-of-Thoughts}{, }Instilling Parallel Reasoning~\cite{InstillingParallelReasoning}, leaf, text width=37.0em
                            ]
                        ]
                        [
                           Mixture of Agents
                            [
                            MoA~\cite{Mixture-of-Agents}{, }MALT~\cite{MALT}{, }SMoA~\cite{SMoA}, leaf, text width=37.0em
                            ]
                        ]
                    ]
                ]
                [
                    Efficiency, ver
                    [
                        Parallel Decoding (\S \ref{Parallel Decoding})
                        [
                        Multiverse~\cite{Multiverse}{, }ParaThinker~\cite{parallel-thinker}{,}APAR~\cite{APAR}{, }PASTA~\cite{LearningtoKeepaPromise}{, }\\ParaDecode~\cite{FastandAccurate}{, }Blockwise~\cite{BlockwiseParallelDecoding}{,}ASPD~\cite{chen2025aspdunlockingadaptiveserialparallel}, leaf, text width=48.5em
                        ]
                    ]
                    [
                        Parallel Function Call (\S \ref{Parallel Function Call})
                        [
                        AsyncLM~\cite{AsynchronousLLMFunctionCalling}{, }LLMCompiler~\cite{AnLLMCompilerforParallelFunctionCalling}{,}ParallelSearch~\cite{ParallelSearch}, leaf, text width=48.5em
                        ]
                    ]
                    [
                        Speculative Decoding (\S \ref{Speculative Decoding})
                        [
                        SSR~\cite{SpeculativeParallelScaling}{, }SpecSearch~\cite{AcceleratingLargeLanguageModelReasoningviaSpeculativeSearch}{, }Speculative Rejection~\cite{FastBest-of-NDecoding}{, }\\Medusa~\cite{MEDUSA}{, }Lookahead~\cite{Lookahead}{,}Jacobi Decoding~\cite{santilli-etal-2023-accelerating}{, }\\Self-Draft~\cite{gao2025multi}{,}Falcon~\cite{gao2025falcon}, leaf, text width=48.5em
                        ]
                    ]   
                ]
            ]
        \end{forest}
    }
    \caption{Taxonomy of Parallel Reasoning}
    \label{taxo}
\end{figure*}
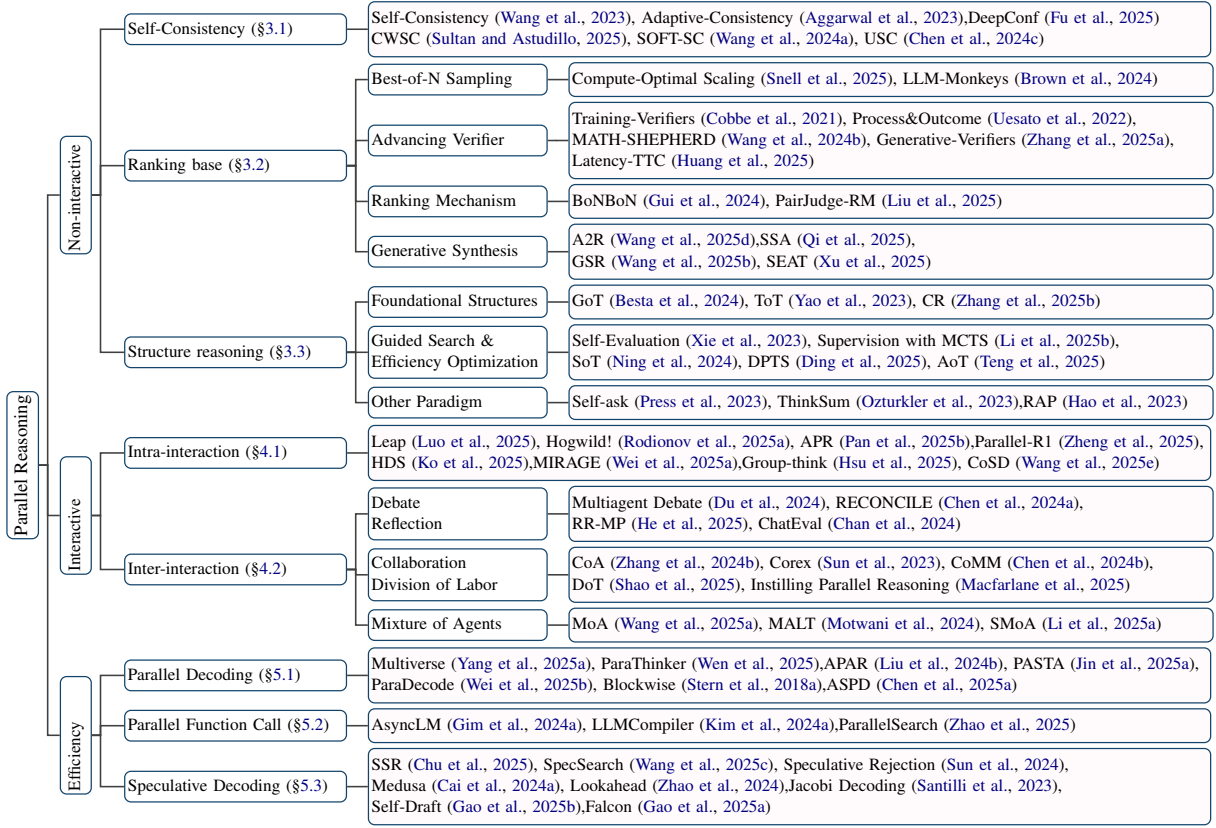
\section{Definition and Formulation}
Here, we provide a formal definition of parallel reasoning, an inference-time paradigm that enhances a language model's reasoning capacity by generating and aggregating multiple concurrent reasoning paths from a single query, thereby producing a more robust final response.


Formally, given an input query $Q$, a language model $M$ produces a final prediction $\Pi(Q)$ through a three-stage pipeline consisting of decomposition, parallel processing, and aggregation:
\[
\Pi(Q) = (A \circ P_M \circ D)(Q)
\]
Here, $D$ is a decomposition operator that maps the input query to a set of sub-inputs, which may represent distinct sub-inputs or identical queries intended to initiate diverse reasoning paths; $P_M$ denotes the parallel application of model $M$ to these inputs; and $A$ is an aggregation operator that synthesizes the intermediate results into the final response.

This formulation captures the essence of parallel reasoning: the ability to process multiple reasoning paths simultaneously and synthesize them into a coherent output. It stands in contrast to traditional sequential reasoning, which processes a single path in a strictly serial manner.

\noindent\textbf{Decomposition} ($D$): Given a query $Q$, the decomposition operator generates a set of $n$ sub-inputs $\mathbf{T} = \{T_1, T_2, \dots, T_n\}$, where each $T_i$ represents a distinct sub-task or problem component:
\[
D(Q) \rightarrow \{T_1, \dots, T_n\}
\]
In the simplest case, $D$ can be an identity mapping that duplicates the original query, i.e., $T_i = Q$ for all $i$, allowing the model to explore diverse reasoning trajectories from the same prompt.

\noindent\textbf{Parallel Processing} ($P_M$): Applies the language model $M$ concurrently to each sub-input $T_i$, producing a set of intermediate results $\mathbf{R} = \{R_1, \dots, R_n\}$:
\[
(R_1, \dots, R_n)= P_M(T_1, \dots, T_n)
\]


\noindent\textbf{Aggregation Operator} ($A$): Combines the set of intermediate results $\mathbf{R} = \{R_1, \dots, R_n\}$ into a single, coherent prediction. 
\[
\Pi(Q) = A(R_1, \dots, R_n)
\]
The behavior of $A$ is characterized by two key properties: its \emph{granularity}—whether aggregation occurs at the sequence level or token level—and the choice of \emph{aggregation function}. 

Based on this definition, parallel reasoning is distinguished from other related inference-time strategies.
Unlike Chain-of-Thought, which decomposes a problem into a strictly sequential series of steps, parallel reasoning processes multiple sub-inputs concurrently. 
A further distinction lies with long-thinking processes, which scale reasoning horizontally by iteratively extending a single reasoning chain through self-reflection and self-refinement. 
In contrast, parallel reasoning scales computation vertically by simultaneously generating multiple distinct reasoning paths.

\section{Non-interactive} \label{non-interactive}
In this section, we introduce \textbf{Non-interactive Parallel Reasoning} as a foundational paradigm wherein multiple candidate reasoning paths are generated without overt communication, followed by a convergence step that aggregates these diverse outputs to produce a single answer.
\[
\Pi(Q) = A(P_M(Q_1, \dots, Q_n))
\]
where $Q_1, \dots, Q_n$ are identical copies of the input query $Q$.
Our examination of this paradigm is structured around three categories.
We begin with the seminal approach of aggregation by self-consistence, which first established the power of consensus in improving accuracy. 
The paradigm then evolved into the now-dominant method of aggregation by ranking—commonly known as Best-of-N sampling—where verifiers or reward models are used to select the optimal answer. 
Finally, we cover a parallel line of research that employs structured exploration to enable the parallel execution of distinct reasoning branches.
\subsection{Self-Consistency} \label{Self-Consistency}
The foundational concept of parallel reasoning stems from Self-Consistency~\citep{Self-Consistency}. 
This approach prompts a model to generate multiple responses and then selects the most common one through a voting process. 
\[
A(R) = \mathop{\mathrm{argmax}}_{a} \sum_{i=1}^{n} \mathbf{I}(E(P_M(Q_i)) = a)
\]
where $E(\cdot)$ is an extraction function that isolates the final answer from an output sequence, $a$ is a candidate answer from the potential answer space, and $I(\cdot)$ is the indicator function.
The underlying principle is that as the number of generated answers increases, the model converges toward a high-confidence solution that more accurately represents its true problem-solving capabilities, whereas a single reasoning path is more susceptible to noise. 

However, this method is computationally expensive and not always practical for general scenarios.
To address these limitations, early work like Adaptive-Consistency~\citep{Let's-Sample-Step-by-Step} introduced a dynamic stopping criterion to optimize the computational budget.
More advanced methods such as DeepConf~\citep{DeepConf} further refine this by leveraging the model's internal confidence scores. 
DeepConf can dynamically terminate low-confidence reasoning paths during generation or use these scores to weight the final votes, drastically reducing computation while boosting accuracy.
Following a similar principle, the Confidence-Weighted Token Set Cover~\citep{CWSC} approach improves token efficiency by pruning intermediate hypotheses based on two indicators: the model's confidence in each path and the lexical diversity across all paths. 
It uses a fast weighted set cover algorithm to select a high-quality and diverse set of reasoning paths, achieving token savings of 10-35\% in many cases.

To generalize the approach to open-ended tasks, Soft Self-Consistency~\citep{Soft-Self-Consistency} replaced majority voting with a "soft" vote based on the model's aggregated generation probabilities, shifting the focus from the most common answer to the one with the highest model-assigned confidence.
In parallel, Universal Self-Consistency~\citep{Universal-self-consistency} uses the LLM itself to judge and select the most consistent response from a set of candidates. 
This evolution from simple consensus to model-based judgment foreshadowed the development of more sophisticated aggregation paradigms.
\subsection{Ranking-Base} \label{ranking}
The Self-Consistency method determines the final answer based on a consensus among multiple generated outputs, typically through majority voting. 
However, its limitation lies in the possibility that the correct answer, while present within the solution space, may not be the most frequent one and is thus never selected.
In contrast, ranking-based methods address this by utilizing an external scoring function—a verifier or reward model(RM)—to evaluate and rank all candidate answers. 
\[
A(R) = E\left(\mathop{\mathrm{argmax}}_{R_i \in \mathbf{R}} V(R_i)\right)
\]
where $V(\cdot)$ is a scoring or verification function that assigns a quality score to an output sequence. 
This is commonly implemented through a Best-of-N (BoN) sampling paradigm, which selects the highest-scoring solution from the generated set. 
Consequently, this approach places immense importance on the quality and sophistication of the selection mechanism, which has become a major area of research in its own right.

The \textbf{BoN} sampling paradigm operates on a simple premise: a model's ability to generate at least one correct solution across $N$ attempts is strictly greater than in a single attempt. 
\citet{Training-Verifiers-to-Solve-Math-Word-Problems} first demonstrated this approach's effectiveness by training a verifier to select correct answers from numerous generated solutions to math problems, significantly boosting performance.
Subsequent work revealed that this improvement follows predictable inference-time scaling laws. Notably, Large Language Monkeys~\citep{Large-Language-Monkeys} found that coverage—the fraction of problems with at least one correct solution—scales consistently with the number of samples $N$. 
Formalizing this further, Compute-Optimal Scaling~\citep{Scaling-LLM-Test-Time-Compute} analyzed the trade-off between model size and sample count for a fixed inference budget. 
Their findings suggest that for large budgets, increasing the sample count $N$ is often a more effective path to improvement than simply scaling up the model.
These studies establish BoN as a fundamental axis for scaling model capabilities, but they also illuminate two parallel challenges: the efficiency of compute allocation and the quality of the verifier.

Addressing the first challenge, the scaling laws reveal a core inefficiency of fixed strategies, which tend to overspend compute on simple queries while under-provisioning for complex ones. 
To solve this, Latency-TTC~\citep{latent-ttc} introduces a query-adaptive approach that dynamically selects the optimal inference method and budget for each query. 
Its key innovation lies in a utility function that uniquely incorporates wall-clock latency—a critical factor for user experience—in addition to token cost. 
Guided by lightweight predictors of accuracy and cost, this adaptive strategy achieves a superior accuracy-efficiency trade-off compared to static approaches.

Complementing these system-level optimizations for compute, a parallel line of research addresses the second challenge by improving the core ranking component itself: the \textbf{verifier}. 
These verifiers, which provide the critical signal for ranking candidates, have evolved into two main approaches: Outcome Reward Models(ORMs) and Process Reward Models(PRMs)~\citep{Solving-math-word-problems-withprocess-and-outcome-based-feedback}. 
An ORM evaluates an entire reasoning chain and assigns a single, holistic score based solely on the final answer's correctness. 
In contrast, a PRM provides more granular feedback by assigning a reward to each intermediate step, a form of supervision that ensures the logical coherence of the process and not just the accuracy of the final output.
The Generative Verifier(GenRM)~\citep{Generative-Verifiers} reframes verification as a next-token prediction task. 
Instead of outputting a score, the verifier generates a "Yes" or "No" token to answer the question, "Is the following solution correct?", with the reward derived from the token's probability. 
This approach also enables GenRM-CoT, where the verifier first generates a rationale explaining its assessment before delivering a verdict. 
Concurrently, the MATH-SHEPHERD~\citep{Math-Shepherd} automates process-wise supervision by labeling a reasoning step as "good" if it has a high potential to lead to the correct final answer. 
Such innovations in verifier training have been spurred to break the data bottleneck created by the expensive, step-by-step human annotations upon which the superiority of traditional PRMs depends.

Given the difficulty of assigning reliable absolute scores, advanced \textbf{ranking mechanisms} often shift to relative judgments. 
PairJudge RM~\citep{PairJudge-RM} trains a model on pairwise comparisons, using a knockout tournament to iteratively filter candidates until a single champion emerges. 
To address the computational expense of such powerful verifiers, BoNBoN~\citep{BoNBoN-Alignment} employs a multi-stage, hierarchical strategy, in which an efficient ranker first filters a large set of candidates into a promising subset, which is then re-ranked by a more powerful and computationally expensive verifier. 
This approach significantly improves efficiency by focusing intensive computation only on the most viable solutions.

A more profound shift moves beyond mere selection to \textbf{generative synthesis}, where the goal is to construct a new, superior solution by integrating insights from all candidates. 
This is often realized through a two-stage "explorer-synthesizer" architecture.
For instance, the Asymmetric Two-Stage Reasoning(A2R) identifies that the synthesizer's capability is the critical driver of performance, which leads to an efficient ``small-to-big" configuration—using a small model for parallel exploration and a larger model for synthesis—that can surpass the performance of a much larger monolithic model at a lower cost.
Similarly, the Sample Set Aggregator(SSA)~\citep{Learning-to-Reason-Across-Parallel} and SEAT~\citep{Adaptive-Termination-for-Multi-round-Parallel} frameworks train dedicated aggregator models with reinforcement learning to review, reconcile, and synthesize a final answer from a set of candidates. 
AGGLM specifically highlights the importance of a balanced training curriculum of easy and hard problems to ensure the model can recover correct answers even when they are not the most frequent.
In contrast to these decoupled approaches, Generative Self-Refinement(GSR)~\citep{learn-refine} trains a single, unified model to perform both roles, first generating diverse candidates and then executing self-refinement on its own outputs. 
This dual capability is instilled through a hybrid training pipeline that jointly optimizes for both direct problem-solving and refinement tasks.
\subsection{Structure Reasoning} \label{structure}
A parallel line of research marks a pivotal shift from the linear progression of Chain-of-Thought to \textbf{structured reasoning}, a paradigm of dynamic problem-solving topologies that shares the conceptual goal of generative synthesis in constructing a final solution by exploring and integrating multiple lines of thought. 
This began with Tree-of-Thoughts~\citep{Tree-of-Thoughts}, which framed reasoning as a tree search, enabling the parallel exploration, self-evaluation, and pruning of multiple solution paths. 
Graph-of-Thoughts~\citep{Graph-of-Thoughts} advanced this concept further by employing a graph structure, which unlocks more sophisticated operations like the aggregation of disparate reasoning lines and the refinement of ideas through feedback loops. Complementing these exploration-based methods, Cumulative Reasoning~\citep{Cumulative-Reasoning-with-Large-Language-Models} offers a distinct, iterative approach that ensures logical robustness by sequentially solving sub-problems and accumulating each validated result into the context for the next step. 
Together, these methods define a new paradigm that moves beyond simple pathfinding to instead construct solutions by synthesizing insights from multiple, potentially incomplete paths, hinting at a future focused on developing a "thought algebra" for formally combining reasoning intermediates.

The vast search spaces and significant inference overhead introduced by these complex structures necessitate two parallel lines of optimization: \textbf{guided search strategies and efficiency enhancements}. 
To navigate the combinatorial explosion of possibilities, Self-Evaluation Guided Beam Search~\citep{Self-Evaluation} uses the model's intrinsic ability to critique candidate steps as a heuristic to prune the search beam. 
In a different paradigm, Process Supervision with Monte Carlo Tree Search~\citep{Enhancing-Reasoning-through-process-supervision} uses the search algorithm not just for guidance but also to generate fine-grained data for progressively enhancing the model's capabilities. Alongside these search improvements, new efficiency optimizations have emerged at distinct layers of abstraction. At the prompting layer, Skeleton-of-Thought~\citep{Skeleton-of-Thought} reduces latency by generating a concise outline before parallel expansion; at the system layer, Dynamic Parallel Tree Search~\citep{Dynamic-Parallel-Tree-Search-for-Efficient-LLM-Reasoning} optimizes execution with advanced KV cache management; and at the theoretical layer, Atom-of-Thoughts~\citep{Atom-of-Thoughts} redefines the process to prevent computational load from accumulating. 
These developments highlight a crucial co-evolution: complex reasoning structures demand sophisticated search and optimization algorithms, which in turn enable and validate more elaborate structures.

A final class of frameworks adopts a different philosophical approach, fundamentally shifting the "locus of intelligence" by redefining the LLM's role. 
Self-ask~\citep{Measuring-and-Narrowing} still treats the LLM as a complete reasoner but compels it to externalize its thought process by breaking down complex problems into explicit sub-questions. 
ThinkSum~\citep{ThinkSum} introduces a sharper division of labor, recasting the LLM as a specialized "System 1" for associative knowledge retrieval while offloading formal "System 2" logical inference to an external model. 
Taking this deconstruction even further, Reasoning via Planning~\citep{Reasoning-with-Language-Model} reframes the task as a classical AI planning problem, repurposing the LLM as a flexible semantic simulator while an external algorithm like MCTS provides the overarching strategic intelligence. 
This conceptual shift—from treating the LLM as a monolithic reasoner to a specialized component—signals the rise of hybrid cognitive architectures and the emerging discipline of "cognitive engineering."

\section{Interactive}
Interactive parallel reasoning is a finer-grained paradigm in which multiple reasoning paths or agents operate in parallel while dynamically exchanging information during inference, rather than generating independently and aggregating only at the end. Formally,

\begin{align*}
\Pi(Q) &= A\!\left(\{R_i^{(T)}\}_{i=1}^n\right), \\[6pt]
R_i^{(t+1)} &= P\!\left(T_i;\,\{R_j^{(t)}\}_{j \neq i}\right).
\end{align*}

where $R_i^{(t)}$ denotes the intermediate output of the $i$-th branch at interaction round $t$. 
Unlike the non-interactive case, each branch is iteratively updated based not only on its own input $T_i$ but also on the intermediate outputs from other branches $\{R_j^{(t)} : j \neq i\}$. 
The process continues for $T$ rounds of interaction before applying the aggregation operator to produce the final answer.

This interaction can take two main forms: \textbf{intra-interaction} (Section~\ref{subsec:intra}), where different reasoning threads within a single model share information and adjust their trajectories during the generation process, 
and \textbf{inter-interaction} (Section~\ref{subsec:inter}), where multiple autonomous models or agents collaborate by exchanging intermediate results or engaging in dialogue. 
By enabling real-time communication among concurrent processes, interactive parallel reasoning supports dynamic error correction, mitigates the influence of early mistakes, and yields more robust and efficient reasoning.

\subsection{Intra-interaction} 
\label{subsec:intra}
\begin{align*}
\Pi(Q) &= A\!\left(\{R_i^{(T)}\}_{i=1}^n\right), \\[6pt]
R_i^{(t+1)} &= P\!\left(T_i, \; \mathcal{S}\!\left(\{R_j^{(t)} : j \neq i\}\right)\right).
\end{align*}
Intra-interaction parallel reasoning allows different reasoning paths within a single model to share intermediate information during generation. 
Formally, each branch is updated by referencing both its own input $T_i$ and the shared information $\mathcal{S}(\cdot)$ aggregated from other branches, and the final answer is obtained through aggregation.

Unlike independent sampling or post-hoc ensembling, intra-interaction parallel reasoning allows the reasoning paths to exchange information rather than remain isolated. Share information during the reasoning process itself so that  each path can be updated in light of what others are producing. This makes the paradigm both parallel, as multiple trajectories are explored simultaneously, and interactive, since intermediate states are exchanged rather than aggregated only at the end. Within this general setting, prior works differ mainly in when and how such exchanges occur. Some methods focus on scheduling decisions, controlling the number of parallel branches to create and when to terminate them. Others introduce staged or step-wise mechanisms that allow paths to periodically exchange summaries or revisions. Still others operate directly at the decoding level, enabling token-by-token visibility and continuous cross-attention across streams. We next review representative works under each of these categories. One group of methods focuses on scheduling-oriented interaction, where the model adaptively controls the number of branches to explore and decides when to terminate them. Entropy-based termination exploits the negative correlation between semantic diversity and correctness to halt exploration once consensus emerges, thereby reducing redundant computation~\citep{Adaptive-Termination-for-Multi-round-Parallel}. Thread-based frameworks capture the same adaptive principle by introducing explicit spawn and join operations: a spawn corresponds to opening additional reasoning branches when more exploration is judged useful, while a join corresponds to merging or terminating branches once verification is sufficient~\citep{Learning-Adaptive-ParallelReasoning}. Extensions to retrieval-augmented reasoning push this idea further by enabling hybrid scheduling: some sub-queries are dispatched in parallel to maximize coverage, while others are routed sequentially to preserve dependency constraints, yielding a dynamic mix of parallel and serial processing within the same task~\citep{hds}. Finally, reinforcement learning has also been applied to directly instill adaptive branching policies into the model itself. Parallel-R1 adopts a progressive curriculum that first familiarises the model with the format of parallel reasoning on simpler problems, then transitions to RL with alternating structural and outcome-based rewards~\citep{parallel-r1}.

A second line of work introduces staged or step-wise interaction, in which different reasoning paths periodically share partial results to incorporate peer feedback. In this setting, LeaP allows each path to generate concise summaries that are routed to other paths under diverse or clustered strategies, helping reduce error propagation and enabling peers to escape locally suboptimal trajectories~\citep{Learning-from-Peers}. Collaborative speculative decoding (CoSD) extends speculative decoding by letting multiple drafts exchange probability and semantic information during generation rather than only at the end, thus turning speculative exploration into a collaborative process~\citep{Speculatethen}. MIRAGE follows a similar staged philosophy in retrieval-augmented reasoning: it decomposes a complex query into multiple sub-questions, launches parallel reasoning chains that iteratively retrieve and update knowledge from a shared workspace, and finally employs a synthesis stage that cross-validates chains and resolves conflicts across their outputs~\citep{MIRAGE}. Together, these methods highlight the role of periodic summarisation, cross-sharing, and staged verification in enabling more robust reasoning than independent parallel sampling. At the decoding level, the most tightly coupled approaches give paths continuous access to each other’s outputs. Group Think realises this by instantiating multiple thinkers whose token streams are mutually visible at every step, producing emergent division of labour and reduced redundancy~\citep{Group-Think}, while Hogwild! Inference achieves a similar effect by enabling concurrent workers to share key–value caches, with RoPE-based adjustments ensuring positional consistency so that each branch can immediately draw on the evolving memory of others~\citep{Hogwild}. Together, these methods illustrate a spectrum of intra-interaction strategies, ranging from adaptive scheduling to periodic summarisation and revision to token-level collaboration, all adhering to the principle that each reasoning path is updated not only from its own trajectory but also from shared information aggregated from its peers.

\subsection{Inter-interaction} 
\label{subsec:inter}
\begin{align*}
\Pi(Q) &= A\!\left(\{R_i^{(T)}\}_{i=1}^k\right), \\[6pt]
R_i^{(t+1)} &= M_i\!\left(Q, \; \mathcal{I}\!\left(\{R_j^{(t)} : j \neq i\}\right)\right).
\end{align*}

Inter-interaction parallel reasoning involves multiple models or agents 
$\{M_1, M_2, \ldots, M_k\}$ that exchange intermediate results during inference. 
Each agent $M_i$ updates its output $R_i^{(t+1)}$ based on the original query $Q$ 
and the interaction function $\mathcal{I}(\cdot)$ aggregating the contributions of other agents. The final prediction $\Pi(Q)$ is obtained by applying the aggregation operator $A$ over all agents' outputs.

Inter-interaction parallel reasoning refers to settings where collaboration occurs across multiple autonomous models or agents, rather than within a single model’s internal processes. In this paradigm, reasoning is distributed among different agents that explicitly exchange intermediate results or engage in conversation. One prominent style is \textbf{debate and reflection}, where multiple agents engage in structured dialogue to reach stronger conclusions than any single model could achieve. Multi-agent debate frameworks show that when independent LLM instances propose answers and then iteratively critique or defend them, the resulting consensus significantly improves both factuality and reasoning quality, while reducing hallucinations~\citep{ImprovingFactualityandReasoning}. Building on this “society of minds” perspective, round-table designs extend the debate to diverse models, incorporating confidence scores and human demonstrations to guide persuasion and convergence, thereby leveraging model heterogeneity as a key driver of performance gains~\citep{ReConcile}. Other work emphasises combining debate with reflection, pairing reactive agents with reflective counterparts in multi-path settings to prevent degeneration of thought and strengthen reasoning robustness, particularly in scientific domains~\citep{EnhancingLLMReasoningwithMulti-Path}. Beyond task-solving, debate has also been applied to evaluation itself: by treating assessment as a group deliberation problem, multi-agent referee teams collaboratively discuss system outputs, leading to judgments that better approximate human evaluation standards~\citep{chateval}.

Beyond dialogue-based approaches, another major direction within inter-interaction parallel reasoning is \textbf{collaboration and division of labour}, which emphasises distributing reasoning workloads across agents or roles in a coordinated manner. A first line of work focuses on task decomposition and management, where complex problems are split into sub-tasks handled by different agents before being synthesized. For example, segmenting long-context inputs among worker agents with a manager responsible for aggregation alleviates context window limitations, while hybrid architectures allocate simpler tasks to lightweight local models and outsource more complex components to larger cloud-based LLMs, thereby reducing costs while maintaining accuracy~\citep{ChainofAgents,Division-of-Thoughts}. A second group emphasises role specialisation and heterogeneous collaboration, where agents are prompted to take on complementary roles or adopt distinct reasoning modes. By incorporating mechanisms such as discussion, review, and retrieval, or by assigning expert roles along different reasoning paths, these frameworks harness diversity in perspectives to improve performance on challenging scientific and reasoning tasks~\citep{Corex,CoMM}. Finally, a third direction explores instilling parallel reasoning capabilities within a single model, where parallel traces distilled from teacher models enable the student model to decompose problems, explore multiple strategies concurrently, and dynamically allocate reasoning depth based on complexity, effectively internalising the benefits of multi-agent division of labor~\citep{InstillingParallelReasoning}. Together, these approaches demonstrate how decomposition, specialisation, and knowledge distillation provide complementary pathways to scale reasoning through collaborative or distributed structures, improving robustness, efficiency, and adaptability.

Building on both debate-based and division-of-labour paradigms, a further strand of inter-interaction research develops \textbf{mixture-of-agent} methods, which focus on harnessing the complementary expertise of multiple agents through structured mixture mechanisms. One approach builds a layered architecture where each agent conditions on the outputs of others in previous layers, allowing collective strengths to be integrated across stages and yielding performance that surpasses even state-of-the-art single models~\citep{Mixture-of-Agents}. A related line of work explores mixture-of-agents in the training paradigm: by dividing reasoning into generation, verification, and refinement, and constructing a multi-agent search tree, cooperative training frameworks enable agents to specialize in different roles and learn from both successful and failed trajectories, ultimately strengthening end-to-end reasoning quality~\citep{MALT}. While these dense multi-agent structures can be powerful, they often introduce efficiency and redundancy challenges; to address this, sparse mixture frameworks restrict interactions among agents through mechanisms such as selective response sharing and early stopping, while also encouraging diversity through distinct role descriptions, thereby achieving performance comparable to dense systems at a fraction of the computational cost~\citep{SMoA}. Taken together, these works highlight how mixture-of-agents, whether through layered inference architectures, cooperative training pipelines, or sparse interaction designs, provide a flexible and scalable means to integrate multiple reasoning agents into a unified system that balances performance, efficiency, and diversity.

\section{Efficiency}
\label{sec:bibtex}

The approaches discussed earlier primarily focus on the horizontal expansion of reasoning, enabling large language models to explore multiple reasoning paths in parallel so as to improve the quality and robustness of their outputs.
In this section, we shift our attention to strategies for enhancing the efficiency of parallel reasoning, which can be categorized by the granularity at which the optimization is applied, including parallel decoding, parallel function calling, and speculative decoding. We formalize these optimization methods as follows:

$$\Pi(Q)=\bigl(A \circ P_{M} \circ D^{(V,\ell)}\bigr)(Q),$$

where $D^{(V,\ell)}$ operates along the vertical (sequential) axis of a single reasoning trajectory, and $\ell \in \{\text{task},\text{call},\text{token}\}$ denotes the granularity of decomposition, which corresponds to parallel decoding, parallel function calling, and speculative decoding, respectively.
At each step, the vertical decomposition operator $D^{(V,\ell)}$ acts on the input task $Q$ and the previous state $s_{t-1}$, partitioning the sequential reasoning process into a set of parallelizable or pipelined sub-units:

$$U_\ell(s_{t-1}) = D^{(V,\ell)}(Q,s_{t-1}) = \{T_1,\ldots,T_m\}.$$

Here, $U_\ell(s_{t-1})$ represents the vertical decomposition result at granularity $\ell$, and each element $T_i$ corresponds to an independent computation unit.
The outputs of these vertically decomposed units are processed by $P_M$ in parallel or pipeline fashion and then integrated by $A$ with order consistency.
In essence, efficiency-oriented methods reinterpret $D$ as a vertical decomposer that reorganizes the temporal granularity of inference to enable higher concurrency within a single reasoning trajectory.

\subsection{Parallel Decoding}
\label{Parallel Decoding}
At the macro level, parallel decoding explores task-level or semantic-level parallelism. By decomposing forthcoming decoding tasks into semantically or structurally independent units and executing them concurrently, these approaches reduce inference latency while preserving generation quality. APAR~\cite{liu2024aparllmsautoparallelautoregressive} fine-tunes LLMs to recognize hierarchical structures (e.g., list items) and fork parallel decoding threads. Multiverse ~\cite{yang2025multiverselanguagemodelssecretly} adopts a Map–Process–Reduce paradigm, training models to decompose tasks, process subtasks in parallel, and merge results losslessly. ParaDecode~\cite{wei2024fast} pipelines token generation by leveraging intermediate-layer confidence to finalize tokens early and begin subsequent computations in parallel, ensuring identical outputs with standard decoding. PASTA~\cite{jin2025learning} introduces a learned annotation language to mark semantically independent text segments, which an interpreter decodes asynchronously, nearly doubling inference speed with negligible impact on output quality. ASPD~\cite{chen2025aspdunlockingadaptiveserialparallel} detects intrinsically parallelizable branches and employs a hybrid engine to seamlessly switch between sequential and parallel modes. Hogwild~\cite{rodionov2025hogwildinferenceparallelllm} enables multiple LLM instances to generate concurrently with shared attention caches, requiring no additional fine-tuning and delivering near-linear scaling.
\subsection{Parallel Function Call}
\label{Parallel Function Call}
At the system level, parallel function calling introduces parallelism through external tool coordination.
Instead of modifying the decoding process itself, it enables LLMs to schedule and execute multiple function calls simultaneously, reducing sequential bottlenecks and improving end-to-end responsiveness. AsyncLM~\cite{gim2024asynchronousllmfunctioncalling} realizes this idea by introducing asynchronous interrupts into generation, allowing the model to continue decoding while functions execute in parallel, and employing scheduling heuristics such as Longest Processing Time first to manage multiple concurrent calls, achieving latency reductions of up to 5.4× without loss of accuracy. LLMCompiler~\cite{kim2024llmcompilerparallelfunction}, in contrast, adopts a compiler-inspired approach: it decomposes multi-call queries into optimized execution plans via a dedicated function calling planner, and dispatches tasks in parallel through a specialized executor.

\subsection{Speculative Decoding}
\label{Speculative Decoding}
At the micro level, speculative decoding achieves token-level parallelism within the autoregressive process by adopting a draft-and-verify paradigm~\cite{leviathan2023fastinferencetransformersspeculative, chen2023acceleratinglargelanguagemodel, xia-etal-2023-speculative}, which accelerates generation and mitigates the sequential bottleneck of token-by-token decoding. The key challenge lies in producing high-quality drafts that can be efficiently verified by the target model. A common approach employs a smaller, faster draft model from the same family as the target LLM to ensure distributional alignment~\cite{leviathan2023fastinferencetransformersspeculative, chen2023acceleratinglargelanguagemodel, xia-etal-2023-speculative}. Acceptance can be further improved through multi-sample generation~\cite{yang2024multicandidatespeculativedecoding} or multiple parallel draft models~\cite{Miao_2024}, though these designs introduce inter-model coordination overhead and retain non-trivial autoregressive costs.

Motivated by the efficiency limitations of conventional speculative decoding, recent work explores more lightweight and self-sufficient alternatives. REST~\cite{he2024restretrievalbasedspeculativedecoding} removes auxiliary models by retrieving contextually relevant tokens from a pre-built corpus and verifying them in parallel, while HD~\cite{cho2025losslessaccelerationlargelanguage} organizes heterogeneous corpora hierarchically to improve draft quality. Building on this idea, CS Drafting~\cite{chen2025cascadespeculativedraftingfaster} and StagedSpec~\cite{spector2023acceleratingllminferencestaged} coordinate multiple draft modules with varying computational costs to reduce latency while maintaining diversity.

Another line of work performs model-internal drafting, where the LLM itself generates and verifies candidate drafts without external auxiliaries. Methods such as Draft \& Verify~\cite{Zhang_2024}, SWIFT~\cite{xia2025swiftontheflyselfspeculativedecoding}, LayerSkip~\cite{Elhoushi_2024}, and Kangaroo~\cite{liu2024kangaroolosslessselfspeculativedecoding} leverage internal representations or adaptive layer skipping to accelerate decoding. Structural extensions further enhance efficiency: Blockwise Decoding~\cite{stern2018blockwiseparalleldecodingdeep}, MEDUSA~\cite{cai2024medusasimplellminference}, Hydra~\cite{ankner2024hydrasequentiallydependentdraftheads}, and the Eagle/Falcon series~\cite{li2025eaglespeculativesamplingrequires, li2024eagle2fasterinferencelanguage, li2025eagle3scalinginferenceacceleration, gao2025falcon} employ multi-head or multi-module architectures to parallelize token generation and verification.

Complementary approaches, including PaSS~\cite{monea2023passparallelspeculativesampling}, SpeDec~\cite{xia2023speculativedecodingexploitingspeculative}, Jacobi Decoding~\cite{santilli-etal-2023-accelerating}, and LADE~\cite{10.5555/3692070.3692631}, achieve acceleration purely through decoding adjustments without retraining, while Self-Draft~\cite{gao2025multi} utilizes the model’s intrinsic robustness to produce context-aware drafts from static corpus priors. Together, these advances demonstrate that efficiency-oriented speculative decoding can be realized through corpus retrieval, model-internal drafting, or architectural parallelization—all aiming to minimize redundancy and maximize throughput.

\section{Application}
Recently, parallel reasoning has emerged as a powerful paradigm for solving complex problems and has been widely adopted by leading AI research teams. 
Beyond pushing the frontiers of performance on grand challenge benchmarks, the principles of parallel reasoning are being applied to enhance the reliability, efficiency, and creativity of large language models in a variety of practical settings.
\paragraph{Solving Grand Challenge Problems}Gemini pioneered an advanced parallel reasoning approach, Deepthink~\citep{gemini-deepthink}, achieving a gold medal-level performance at the International Mathematical Olympiad (IMO) and demonstrating the striking potential of these techniques. 
Following this, other frontier models, such as Grok4\citep{Grok4-heavy} and Claude4~\citep{claud4-heavy}, have integrated similar parallel technologies to significantly boost their performance on mathematical and coding tasks. 
More recently, models like SEED1.6~\citep{seed1.6-thinking} and Qwen3 ~\citep{Qwen3-thinking} have likewise employed parallel decoding techniques to achieve remarkable breakthroughs in competitive contests such as AIME, even approaching perfect scores. 
Beyond complex task-solving, these parallel techniques are also being applied in advanced research systems. For instance, Anthropic utilizes prompt engineering to enhance its multi-agent parallel function calling system~\citep{claud-research}, enabling it to tackle real-world research questions with high efficiency and low latency.
\paragraph{Enhancing Reliability and Factuality}A significant application of parallel reasoning is in mitigating hallucinations and improving the factual accuracy of LLM outputs. 
By generating multiple, diverse reasoning paths, models can identify and discard factual inconsistencies or logical fallacies that might arise in a single-path generation. 
This is exemplified in multi-agent~\citep{ImprovingFactualityandReasoning,ReConcile,Self-Consistency} frameworks, where multiple models propose, critique, and refine answers to yield a final consensus that is significantly more factual and robust.
\paragraph{Accelerating Complex Workflows}Through parallel function calling, models can schedule and run multiple functions concurrently instead of waiting for them to execute sequentially. 
This dramatically reduces bottlenecks in real-world applications requiring data retrieval or external computations. 
Furthermore, models can utilize optimized execution plans~\citep{AsynchronousLLMFunctionCalling,AnLLMCompilerforParallelFunctionCalling} that dispatch tool calls in parallel, leading to significant reductions in end-to-end latency.
At a more fundamental level, the philosophy of parallelism is also applied to accelerate the core token-generation process itself. Techniques such as speculative and parallel decoding~\citep{AcceleratingLargeLanguageModelReasoningviaSpeculativeSearch,FastBest-of-NDecoding,SpeculativeParallelScaling} are designed to parallelize parts of the autoregressive sequence generation, improving system throughput and responsiveness.
\paragraph{Augmenting Creativity and Open-ended Generation}For tasks that do not have a single correct answer, such as brainstorming, creative writing, or code design, parallel reasoning is a powerful tool for exploring a wider solution space and generating more diverse outputs.
By building out a structure of interconnected ideas, methods like ~\citep{Tree-of-Thoughts,Graph-of-Thoughts} allow the model to explore many creative avenues before converging on a final product. Additionally, other approaches~\citep{A2R} generate a diverse set of initial drafts in parallel and then integrate the best elements from all candidates to create a superior final output. 
This process allows for the generation of more novel and high-quality content than any single path could produce.
\section{Challenge and Future Work}
\subsection{Challenge}
\paragraph{Performance Constrain}While current parallel reasoning methods show performance surpassing simple voting baselines, several key challenges remain.
(1) These methods are fundamentally constrained by a Pass@k performance upper bound. 
Although effective at selecting the best solution from a set of generated candidates, they struggle to innovate a novel answer that surpasses the quality of the best available option in the pool.
(2) The performance gains from parallel reasoning exhibit a trend of diminishing returns. 
As the number of parallel samples $N$ increases, the incremental improvement in accuracy decays. 
This suggests that current aggregation strategies do not fully utilize all the useful information spread across the candidates, a limitation that may be rooted in the model's intrinsic reasoning capabilities.
(3) Emerging methods that integrate decomposition and aggregation into a single, adaptive model offer improvements in latency and efficiency. However, their performance is often only comparable to that of simple majority voting, and they have not yet been scaled effectively to achieve a significant performance breakthrough.
\paragraph{Optimization Problem}Current parallel reasoning frameworks also face two primary optimization challenges.
(1) Many two-stage or multi-agent architectures are optimized in a disjointed manner, where different components are trained separately. This approach lacks a unified, end-to-end paradigm that would allow feedback from the aggregation stage to supervise and improve the initial generation stage, hindering holistic performance.
(2) The aggregation stage itself presents significant hurdles. Firstly, when presented with multiple reasoning paths as context, aggregator models often produce a concise summary instead of a novel, comprehensive reasoning trajectory. 
Secondly, training an aggregator using outputs from these independently generated paths introduces a complex off-policy optimization problem, which can destabilize large-scale reinforcement learning and make the approach difficult to scale effectively.
\subsection{Future Work}
\paragraph{Multimodal}The potential of parallel reasoning extends beyond mathematical and coding domains, showing significant promise for boosting performance in multimodal reasoning. 
This paradigm is applicable not only to text-centric multimodal tasks~\citep{GEM,mqa}, such as multimodal question answering and named-entity recognition , but also holds great potential for image-based reasoning~\citep{ref-image}. 
In this context, a model could generate multiple image variations in parallel and then perform complex reasoning based on the combination of these images, opening new avenues for creative and analytical tasks.
\paragraph{End to End Optimization \& Scale up}Future work in parallel reasoning faces significant challenges in training and optimization. 
A primary goal is to develop a unified, end-to-end training paradigm for the entire pipeline, from decomposition to aggregation, which would require a fine-grained, clear, and scalable reward signal to enable holistic optimization. Furthermore, large-scale experimentation is needed to validate that parallel reasoning can function as a continuously improving learning process, rather than merely acting as a Pass@k approximator that selects from a fixed-quality pool of answers. Finally, developing a suitable paradigm for stable, on-policy-like reinforcement learning is crucial. 
Such stability would enable scaling to a much larger number of parallel samples, reducing reliance on long sequential computations and offering a clear path toward improved inference scalability and lower latency.
\section{Discussion and Conclusion}
\subsection{Why does parallel reasoning work?}
\paragraph{DFS vs BFS} Sequential reasoning methods are analogous to a depth-first search (DFS). 
They generate solutions along a single path via iterative refinement, which risks getting trapped in a local minimum. 
In contrast, parallel reasoning resembles a breadth-first search (BFS), exploring multiple potential solutions simultaneously then aggregating them into a robust answer. 
This also indicates an interesting phenomenon: why simple repeated sampling methods can surpass the 'plan-process-summary' parallel method, which can be constrained by the quality of its initial plan.
\paragraph{Scaling Components} Following the evolution of parallel reasoning, repeated sampling has become the most common paradigm for boosting test-time computation. 
A key trend in this area is the evolution of aggregation methods, which have shifted from simple self-consistency (voting) to more sophisticated ranking-based methods (scoring), and ultimately to generative approaches. 
This evolution marks a significant difference in the computational scaling of the aggregation stage. 
While voting methods rely on simple, rule-based selection of the most common answer, ranking methods like ORM/PRM require a full model forward pass to score each candidate, similar to reward models in modern reinforcement learning. 
Generative methods take this even further, re-reasoning over all parallel outputs to construct a final answer, which represents another level of scaling test-time compute. 
Thus, parallel scaling increases computation not only by decomposing a problem into several tasks but also by generating multiple solutions in parallel. 
This scaling of the aggregator's computation also boosts final performance, which can be further improved by scaling the main model or the reward model's inference time. 
This progression highlights a central theme: allocating more computation to all components at test-time can lead to significantly better performance.
\paragraph{Tuning vs In-Context} Sequential reasoning methods, like R1, operate by first rolling out multiple reasoning trajectories. 
These trajectories are then rewarded based on their final answers. Finally, paths are used to fine-tune the model’s parameters through a mechanism that reinforces positive outcomes and penalizes negative ones.
In contrast, native parallel reasoning treats these rollout trajectories as a unified context by concatenating them. 
The model then performs its reasoning based on this combined input, effectively transforming a fine-tuning task into an in-context reasoning task. 
This approach also allows the model to be trained on how to best utilize this parallel information, thereby enhancing its capabilities.
\subsection{Conclusion}
This survey provides a comprehensive overview of parallel reasoning, a pivotal inference paradigm that enhances robustness by shifting from single-path sequential processing to a breadth-first exploration of a model's capabilities. 
Through a unified formal framework, we categorize the field and trace its evolution from early consensus-based methods like self-consistency to sophisticated ranking models and dynamic multi-agent collaboration. 
Despite this progress, the field faces significant challenges, including performance constraints imposed by the initial candidate pool and the lack of unified, end-to-end optimization. 
Ultimately, parallel reasoning represents a fundamental shift toward more deliberate and scalable AI, and we hope this survey catalyzes further innovation in this promising domain.
\bibliography{custom, decode}

\appendix

\end{document}